\documentclass[11pt,a4paper]{article}
\usepackage[hyperref]{acl2021}
\usepackage{times}
\usepackage{latexsym}
\usepackage{pifont}

\newcommand{\COMMENT}[1]{}

\usepackage{microtype}

\aclfinalcopy 
\def\aclpaperid{2635} 

\usepackage{xcolor, colortbl}

\usepackage{multirow}

\usepackage{graphicx}
\graphicspath{ {./images/} }

\usepackage{soul}

\title{{C}ascade \textit{versus} {D}irect Speech Translation: \\ {D}o the Differences Still Make a Difference?\\
}

\author{Luisa Bentivogli\textsuperscript{1}, Mauro Cettolo\textsuperscript{1}, Marco Gaido\textsuperscript{1,2},\\ \textbf{Alina Karakanta\textsuperscript{1,2}, Alberto Martinelli\textsuperscript{2}\thanks{  \textcolor{white}{$\ast$} The work of Alberto Martinelli was carried out during an internship at Fondazione Bruno Kessler.} , Matteo Negri\textsuperscript{1}, Marco Turchi\textsuperscript{1}} \\
  \textsuperscript{1}Fondazione Bruno Kessler \\
  \textsuperscript{2}University of Trento \\
  \texttt{\{bentivo,cettolo,mgaido,akarakanta,negri,turchi\}@fbk.eu}
  }

\date{}

\begin{document}
\maketitle
\begin{abstract}
Five years after the first published proofs of concept, \textit{direct} approaches to speech translation (ST)
are now competing with  traditional \textit{cascade} solutions. In light of this steady progress, can we claim that the performance gap between the two is closed?  Starting from this question, we present  a systematic  comparison between state-of-the-art systems representative of  the two paradigms. Focusing on three language directions (English--German/Italian/Spanish), we conduct automatic and manual evaluations, exploiting high-quality professional post-edits and annotations. 
Our multi-faceted analysis  on one of the few publicly available ST benchmarks attests for the first time  that: \textit{i)} the gap between the two paradigms is now closed, and \textit{ii)} the subtle differences observed in their behavior are not sufficient
for humans neither to distinguish them nor to prefer one over the other. 
\end{abstract}

\section{Introduction}
Speech translation (ST) is the task of automatically translating  a speech signal in a given 
language into a text in another 
language. Research on ST dates back to the late eighties and its evolution followed the development of the closely related fields of speech recognition (ASR) 
and machine translation 
(MT)  that, since the very beginning, 
provided the main pillars for building the so-called \textit{cascade} architectures. With the advent of deep learning, the neural networks widely used in ASR and MT 
have been adapted to develop  a new \textit{direct} 
ST 
paradigm. This 
approach aims to overcome known limitations of the cascade one (e.g. architectural complexity, error propagation)  with a single encoder-decoder architecture that directly translates the source signal bypassing intermediate 
representations.

Until now, the consolidated underlying technologies and the richness of available data have upheld the supremacy of cascade solutions in industrial applications. However, architectural simplicity, reduced information loss and 
error propagation are the ace up the sleeve of the direct approach, which has rapidly gained popularity within the research community in spite of the critical bottleneck represented by data paucity.

Within a few years after the first proofs of concept
\citep{berard_2016,weiss2017sequence}, 
the performance gap between the two paradigms has gradually decreased. This trend is mirrored by the findings of
the International Workshop on Spoken Language Translation (IWSLT),\footnote{\url{http://iwslt.org}}
a yearly evaluation campaign 
where direct systems made their first appearance in 2018. On English-German, for instance, the  BLEU difference between the best cascade and direct models  dropped from 7.4 points in 2018 \citep{jan2018iwslt} to 1.6 points in 2019 \citep{jan2019iwslt}. In 2020, 
participants were  allowed to choose between processing a pre-segmented version of the test set or the one produced by their own segmentation algorithm. As reported in \citep{ansari-etal-2020-findings}, the distance between the two paradigms further decreased to 1.0 BLEU point in the first condition and, for the first time, it was slightly  in favor of the best direct model in the second condition, with a small but nonetheless meaningful 0.24 difference.

So, quoting ~\citet{ansari-etal-2020-findings},  is the cascade solution still the dominant technology in ST?
Has the direct approach
closed the huge initial performance gap? 
Are there systematic differences in the outputs of the two technologies? Are they distinguishable? Answering these questions is more than running an evaluation exercise. 
It implies  pushing research towards a deeper investigation of direct ST, finding a path towards its wider adoption in industrial settings and motivating higher engagement  in data exploitation and 
resource creation to train the data-hungry end-to-end neural systems.

For all these reasons, while \citet{ansari-etal-2020-findings} were 
cautious in drawing firm conclusions, in this paper we delve deeper into the problem with the  first  thorough comparison between the two paradigms. Working on three language directions
(en--de/es/it),
we  train state-of-the-art cascade and direct models  ($\S$\ref{sec:setting}), running them on  test data drawn from the MuST-C corpus~\citep{MuST-Cjournal}.
 
Systems' behavior is analysed from different perspectives, by exploiting high-quality post-edits and annotations by professionals.
After discussing overall systems' performance
($\S$\ref{sec:results}), we move to more  fine-grained automatic and manual analyses covering two main aspects: the relation between systems' performance and specific characteristics of the input audio ($\S$\ref{sec:audio}), and the possible differences in terms of lexical, morphological  and word ordering errors ($\S$\ref{sec:errors}).
We finally explore whether, due to latent characteristics overlooked by all previous investigations, the output of cascade and direct systems can be distinguished either by a human or by an automatic classifier  ($\S$\ref{sec:classif}).
Together with a comparative study
attesting the parity of the two paradigms on our test data, another contribution of this paper is the release of the  manual post-edits that rendered our investigation possible. The data is available at: \url{https://ict.fbk.eu/mustc-post-edits}.

\section{Background}
\label{sec:relwork}

\textbf{Cascade ST.}
By concatenating ASR and MT components \citep{StentifordSteer88,waibel1991janus}, cascade ST architectures represent an intuitive solution to achieve reasonable performance and high adaptability across languages and domains. At the same time, however, they suffer from well-known problems related to the concatenation of multiple systems.
First, they require 
\textit{ad-hoc} training and maintenance procedures for the ASR and MT modules;
second, they suffer from 
error propagation 
and from the loss of speech information 
(e.g. prosody)
that might be useful to improve 
final translations.
Research  has focused on mitigating  error propagation by:
\textit{i)} feeding the MT system with ASR data structures (e.g. ASR n-best, lattices or confusion networks) which are more informative than the   1-best output  \citep{lavie1996multi,matusov2005phrase,bertoldi2005new,beck2019neural,sperber2019self}, and  \textit{ii)} making the MT robust to ASR errors, for instance by  training it on parallel data incorporating real or emulated ASR errors as in \citep{peitz2012spoken,ruiz2015adapting,sperber2017toward,cheng2019breaking,digangi2019robust}.
Although the former solutions 
are effective to some extent, state-of-the-art cascade architectures \citep{pham2019iwslt,bahar2020start} prefer the latter, as they
are simpler to implement and 
maintain.

\noindent
\textbf{Direct ST.}
To overcome the limitations of cascade 
models, \citet{berard_2016} and \citet{weiss2017sequence} proposed the first direct solutions bypassing intermediate representations by means of encoder-decoder architectures based on recurrent neural networks.
Currently, more effective solutions \citep{potapczyk-przybysz-2020-srpols,bahar2020start,gaido-etal-2020-end} rely on ST-oriented adaptations of 
Transformer \citep{transformer} integrating the encoder with: 
\textit{i)} 
convolutional layers
to reduce input length,
and 
\textit{ii)}  
penalties biasing 
attention to local context in the encoder self-attention layers \citep{povey-2018-time-restricted,Sperber2018,digangi2019adapting}. 
Though effective, these architectures have to confront with training data paucity, a critical bottleneck for neural solutions.
The problem has been mainly tackled  with data augmentation and knowledge transfer techniques. 
Data augmentation 
consists in producing artificial training 
corpora
by altering 
existing datasets
or by generating 
(\textit{audio}, \textit{translation}) pairs through speech synthesis or MT \citep{bahar-2019-specaugment,nguyen2019improving,ko2015augmentation,jia2018leveraging}.
Knowledge transfer \citep{Gutstein-et-al-knowledge}   consists in 
passing (here to ST) the knowledge learnt by a neural network trained on closely related tasks (here, ASR and MT). 
Existing ASR models  have been used for encoder pre-training \citep{berard2018end,bansal-etal-2019-pre,bahar2019comparative} and multi-task learning \citep{weiss2017sequence,anastasopoulos-2018-multitask,indurthi2019metalearning}. 
Existing neural MT models have been used for decoder pre-training \citep{bahar2019comparative,inaguma-etal-2020-espnet}, joint learning \citep{indurthi2019metalearning,liu2020bridging} and knowledge distillation \citep{liu2019endtoend}.

\noindent
\textbf{Previous comparisons.}
Most of the works on direct ST also  evaluate the proposed solutions  against a cascade counterpart. 
The conclusions, however, are discordant.
Looking at recent works, \citet{pino-et-al-2019-harnessing} show similar scores, \citet{indurthi2019metalearning} report higher results for their direct model, while \citet{inaguma-etal-2020-espnet}
end up with the opposite finding.
The main problems of these comparisons are that:
\textit{i)} not all the architectures are equally 
optimized, 
\textit{ii)} 
for the sake of fairness
in terms of training data, cascade systems are restricted to unrealistic settings with small training corpora that penalize their performance, and \textit{iii)} 
evaluation always relies only on automatic metrics 
computed on single references.
The IWSLT campaigns \citep{iwsltprec_2019,ansari-etal-2020-findings} 
set up a shared evaluation framework where systems built on a large set of training data are optimized to achieve the best performance,  independently from the underlying architecture.
In the last round, direct models approached, and in one case \citep{potapczyk-przybysz-2020-srpols}
outperformed, the cascade 
ones.
However, the evaluation was run only on one language pair, 
by solely relying on automatic metrics and single references.
In this paper, we overcome these limitations by comparing the
two paradigms on three language pairs, using different metrics, multiple references (including professional post-edits) as well as fine-grained 
automatic and 
manual analysis procedures.

\section{Experimental Setting}
\label{sec:setting}

\subsection{ST Systems}
\label{ssec:systems}

To maximize the cross-language comparability of our analyses, we built the cascade and direct  ST systems for en--de/es/it 
with the same core technology, based on Transformer.  
Their good quality is attested by the comparison with the winning system at the  IWSLT-20 offline ST 
task~\citep{bahar2020start},\footnote{In the pre-segmented 
data condition \citep{ansari-etal-2020-findings}.} 
which consists of 
an ensemble of two cascade models scoring 28.8 BLEU on the  en-de portion of the 
MuST-C 
\textit{Common}
test set. 
On the same 
data, our cascade and direct models 
achieve similar  BLEU scores,
respectively 28.9 and 29.1 (see Table~\ref{tab:overall}).\footnote{Also the ASR performance of our cascade solution (10.2 WER 
on MuST-C \textit{Common}) is in line with the results obtained by \citet{bahar2020start} for their best ASR model.}
%
On en-es and en-it, identical architectures perform similarly or better (up to 32.9 BLEU on en-es). Although BLEU scores are not strictly comparable across languages, we can safely consider all our models as  state-of-the-art.


For the sake of reproducibility, we provide complete details about data, architectures and training setup in Appendix \ref{appsec:systems}.

\subsection{Evaluation Methodology}
\label{ssec:method}

\paragraph{Data.}
Our evaluation data is drawn from the TED-based MuST-C corpus~\citep{MuST-Cjournal},  
the largest
freely available multilingual corpus for ST.
It covers 14 language directions, with English audio segments automatically aligned with their
corresponding manual transcriptions and translations. 
The en--de/es/it MuST-C \textit{Common} test sets contain the same 27 TED talks, for 
a total of around 2,500 segments largely overlapping across languages.\footnote{MuST-C \textit{Common} segments can vary across languages due to the automatic procedures of segmentation,
audio-text alignment and filtering that were applied to the talks.}
For all the three language pairs,
we selected  subsets of MuST-C \textit{Common} containing
the same English audio portions from each talk, in order to obtain representative groups of contiguous segments that are comparable across languages. 
Furthermore, to ensure high data quality, we manually checked the selected samples and kept only 
those segments for which the \textit{audio-transcript-translation} alignment was correct. 
Each of the three resulting test sets -- henceforth \textit{PE-sets} -- is composed of 550 segments, corresponding to about 10,000 English source words.

\paragraph{Post-editing.} 

A key element of our multi-faceted analysis is human post-editing (PE), which consists in manually correcting systems' output according to the input (the source audio in our case). In PE-based evaluation, the original output is compared against its post-edited version using distance-based metrics like TER \cite{Snover:06}.
This allows for counting only the true errors made by a system, without penalising differences due to linguistic variation as it happens when exploiting independent references.
This makes  PE-based evaluation one of the most prominent  methodologies  used  for translation quality  assessment~\citep{Snover:06,snover2009fluency,Denkowski:2010,Cettolo:2013,Bojar2015FindingsOT,C16-1294,BentivogliIWSLT:2018}.
%

To collect the post-edits for our study, we strictly followed the methodology of the IWSLT 2013-2017 evaluation campaigns~\citep{Cettolo:2013}, which offered us a consolidated framework and best practices to draw upon.
%
%
Our cascade and direct systems were both run on the PE-sets to be post-edited.
%
To guarantee high quality post-edits, for each language we hired two professional translators with experience in  subtitling and post-editing. 
Moreover, in order to cope with translators' variability (i.e. more/less aggressive editing strategies),
the outputs of the two ST systems were randomly assigned
ensuring that each translator worked on all the 550 segments, post-editing an equal number of outputs from both systems.
%
%
The 
task was performed with a 
CAT tool\footnote{\url{www.matecat.com}} that displays the manual transcript of the audio together with the ST output to be edited. However, since ST systems take as input an audio signal, we also provided translators with the audio file of each segment, asking them to post-edit strictly according to it.\footnote{The \textit{ad-hoc} ST PE guidelines given to translators are included in 
Appendix \ref{appsec:pe_guidelines}.}
For each language pair, the final PE-set used in our study consists of the 550 MuST-C original \textit{audio-transcript-translation} triplets plus two additional sets of reference translations, i.e. the post-edited versions of the two systems' outputs.

\paragraph{Analyses.}
%
%
%
The collected post-edits are exploited to assess overall systems' performance ($\S$\ref{sec:results}) as well as to carry out deeper quantitative and qualitative analyses
aimed to shed light on possible systematic differences in systems' behavior 
($\S$\ref{ssec:auto-audio} and $\S$\ref{subsec:auto-errors}).
Focusing on 
specific aspects of the ST problem, the inquiry is also performed  
by means of manual annotation of systems' outputs
($\S$\ref{subsec:man-audio}, $\S$\ref{subsec:man-errors} and $\S$\ref{subsec:humanclass}).
%
Due to the linguistic nature of this task, centred on fine-grained aspects requiring a variety of skills in both evaluation and ST technology, for such analyses we relied on three researchers in translation technology -- one per language pair -- with a strong background in linguistics, excellent knowledge of the addressed languages (C2 or native), as well as strong expertise in systems' evaluation.

\section{Overall Systems' Performance}
\label{sec:results}
We compute overall performance results both on the PE-sets 
and on the MuST-C \textit{Common} test sets.
Our primary evaluation 
is based on the collected post-edits.
%
%
%
%
%
We consider two TER-based\footnote{\url{www.cs.umd.edu/~snover/tercom}} 
metrics: 
\textit{i)}  human-targeted TER (HTER) computed between the automatic translation and its human post-edited version,
and 
\textit{ii)} multi-reference TER (mTER) computed against  the closest reference among the three available ones
(two
post-edits and the 
official
reference from MuST-C).
The latter metric better accounts for post-editors' variability, making the evaluation more reliable and informative. 
%
%
%
For the sake of completeness, in Table \ref{tab:overall} we also report  SacreBLEU\footnote{BLEU+c.mixed+\#.1+s.exp+tok.13a+v.1.4.3}
\cite{post-2018-a-call} and TER scores
computed only on the official MuST-C \textit{Common} references.

\begin{table}[ht]
 \centering  
 \footnotesize 
 \setlength{\tabcolsep}{3pt}
\begin{tabular}{l|l|l|l|l|l||l|l}
&& \multicolumn{4}{c||}{PE Set} &   \multicolumn{2}{c}{M. \textit{Common}} \\
\cline{2-8}
      &  & \cellcolor{gray!20}HTER & \cellcolor{gray!20}mTER  & BLEU   & TER   & BLEU  & TER\\
           \hline

\multirow{2}{*}{de}  & C   & \cellcolor{gray!20}28.65 & \cellcolor{gray!20}24.41& 28.96 & 53.23  &28.86 & 53.93\\
                    & D  & \cellcolor{gray!20}30.22 & \cellcolor{gray!20}25.60 & 28.46 & 52.56  & 29.05 & \textbf{52.77}$^\ast$\\
 \hline\hline
           
\multirow{2}{*}{es}&  C  & \cellcolor{gray!20}29.96 & \cellcolor{gray!20}25.30 & \textbf{34.05}$^\ast$ & 50.75   & \textbf{32.93}$^\ast$ & \textbf{53.21}$^\ast$ \\
& D  & \cellcolor{gray!20}\textbf{28.19}$^\ast$ & \cellcolor{gray!20}\textbf{24.02}$^\ast$ & 32.17 & 51.08   & 31.98 & 54.00\\

 \hline\hline

\multirow{2}{*}{it} & C  & \cellcolor{gray!20}25.69 & \cellcolor{gray!20}23.29 & \textbf{30.04}$^\ast$ & 54.01   & 28.56 & 56.29\\
& D  & \cellcolor{gray!20}26.14  & \cellcolor{gray!20}23.26 & 28.81  & 54.06  & 28.56 & \textbf{55.35}$^\ast$\\

 \hline
\end{tabular}
    \caption{
    Performance of
    (C)ascade and (D)irect 
    systems on the PE-sets and  MuST-C \textit{Common} 
    test 
    sets.
    Statistically significant differences ($^\ast$) are computed with Paired Bootstrap Resampling~\citep{koehn-2004-statistical}.
    } 
    \label{tab:overall}
\end{table}

A bird's-eye view of the results shows that, in  more than half of the cases, 
performance differences between cascade and direct systems are not statistically significant. When they 
are, the raw count of wins for the two approaches is the same (4), attesting their substantial parity.

Looking at our primary metrics 
(HTER and mTER),
systems are on par on en-it and en-de, while for en-es the direct approach significantly outperforms the cascade one. This difference, however, does not emerge with the other metrics.
%
%
%
%
%
Indeed, BLEU and TER scores computed against the official  references are less coherent across metrics and test sets. 
%
For instance, on the en-it PE-set the cascade system significantly outperforms the direct one in terms of BLEU score, 
while TER shows the opposite on MuST-C \textit{Common}.
%
%
%
%
%
Interestingly,
the scores obtained using 
independent references can 
also disagree
with those computed with post-edits. This is the case of en-es, where significant HTER and mTER reductions attest the superiority of the direct system, while most  BLEU and TER scores  are still in favor of the cascade.
%


On  the  one hand, primary 
evaluation scores suggest
that the rapidly advancing direct technology has eventually reached the traditional cascaded approach.
On the other, the highlighted incongruities 
confirm
%
widespread concerns about the reliability of fully automatic metrics -- based on independent references -- to properly evaluate neural systems \cite{way2018quality}.
This calls for 
 deeper
quantitative and qualitative analyses. Those presented in the next sections  investigate performance differences focusing on two main aspects: the impact of specific input audio properties ($\S$\ref{sec:audio}), and  the linguistic errors made by the systems ($\S$\ref{sec:errors}).









\section{ST Quality and Audio Properties}
\label{sec:audio}

\subsection{Automatic Analysis}
\label{ssec:auto-audio}
The two ST approaches handle the input audio differently: the cascade one by means of a dedicated ASR component that produces intermediate transcripts; the direct one by extracting all the relevant information to translate in an end-to-end fashion. Is it therefore possible that some audio properties have different impact on their results? Overall performance being equal, answering this question would help to understand if one approach is preferable over the other under specific audio conditions. 

Among other possible 
factors (e.g. noise, recording conditions, overlapping speakers) we tried to shed light on this aspect by focusing on two common factors: audio duration and speech rate. 
%
%
To this aim, we grouped the sentences in the PE-set according to the 
sentence-wise HTER percentage difference
-- i.e.
the difference between the cascade and direct HTER scores divided by their average.
%

The threshold  for considering 
performance differences 
as significant was set to 10\%.
%
The resulting groups 
contain sentences where: \textit{i)} 
cascade is significantly better than  
direct,
\textit{ii)} 
direct is significantly better than  
cascade,
\textit{iii)}
the difference between the two is not 
significant,
and \textit{iv)} both systems have 
HTER=0.
For each 
group, we calculated the average audio duration and the corresponding speech rate in terms of phonemes\footnote{
Obtained by processing the transcripts with eSpeak (\url{espeak.sourceforge.net}).} per second. 

Results are 
shown
in Table~\ref{tab:diff}, where -- for the sake of 
completeness -- also the length of the reference audio transcript is given, together with the average HTER of the systems.

\begin{table}[ht]
    \tabcolsep4pt
    \centering 
    \small
    \begin{tabular}{l|l|c|c|c|c||c|c}
    
     \multicolumn{2}{c|}{} & \rotatebox[origin=b]{90}{\#sentences} & \rotatebox[origin=b]{90}{\shortstack{audio duration \\ (seconds)}} & \rotatebox[origin=b]{90}{\shortstack{speech rate \\ (phonemes/s)}} & \rotatebox[origin=b]{90}{\shortstack{ref transcr length\\  (\#words)}} & \rotatebox[origin=b]{90}{C HTER}  & \rotatebox[origin=b]{90}{D HTER}  \\
         \hline
\multirow{4}{*}{de}   & C better  &  240   &  6.15  &  14.43 & 19.75	 & 16.30  & 40.53\\
                      & D better   &  191   &  6.00 &	14.52  & 18.88	   &  44.85 & 17.89\\
                      & No Diff         &  45   &  6.68  &	 14.31 & 22.07	   & 40.74  & 40.18\\
                      & HTER 0          &  74   &  2.71  &	15.53  & 9.64	        & 0      & 0 \\
    \hline \hline
\multirow{4}{*}{es}   & C better  &  215   &  5.92  &  12.20 & 19.52	 & 16.09  & 38.76\\
                      &  D better  &  234   &  6.28  &  12.09 & 20.39	 &  46.45 & 21.14\\
                      & No Diff         &  54    &  6.47  &	12.01 & 20.26	 &  40.22 & 40.37\\
                      & HTER 0          &  47    &  3.09	 & 13.14 & 10.23  &  0     &  0    \\
    \hline \hline    
\multirow{4}{*}{it}   & C better   &  231   &  6.03     & 12.31  & 19.41  &  14.82 & 36.40 \\
                        & D better  &  212   &  6.06     &	12.21  & 19.33 	&  37.65 & 15.80 \\
                        & No Diff        &  55    &  6.93     &	11.94  & 21.73	&  35.39 & 35.37  \\
                        & HTER 0         &  52    &  2.96     &	12.68  & 10.33	&  0     & 0      \\
    \hline  
    \end{tabular}
    \caption{Comparison of (C)ascade and (D)irect performance based on different audio properties.} 
    \label{tab:diff}
\end{table}

As we can see, results are coherent across languages: audio duration and speech rate averages do not differ, neither when one system performs significantly better than the other, nor when the HTER differences are not significant.
We can hence conclude that, if audio duration and speech rate have any influence on systems' performance, 
our analysis does not highlight specific conditions 
that are more favorable to one approach 
than to the other. Both are equally robust with respect to 
the audio properties here considered.
\subsection{Manual Analysis}
\label{subsec:man-audio}
Handling the input audio differently, the two 
approaches have inherent strengths and weaknesses. In particular, although suffering from the well-known scarcity of  sizeable training corpora,  direct solutions come with the promise \cite{sperber-paulik-2020-speech} of: \textit{i)} higher robustness to error propagation, and \textit{ii)} reduced loss of speech information 
(e.g. prosody).
Our next qualitative analysis tries to delve into these 
aspects by looking at audio understanding and prosody issues.

\paragraph{Audio understanding.} Errors due to wrong 
audio understanding are easy to identify for cascade systems -- since they are evident in the intermediate ASR transcripts -- but harder to spot for direct systems, whose internal representations are by far less accessible.
%
In this case, 
errors can still be identified
in mistranslations corresponding to words
which are
phonetically similar
to parts of the input audio -- e.g. \textit{nice voice} 
mistranslated in German as \textit{nette Jungen} (\textit{nice boys}).
To spot such errors, our annotators carefully inspected the PE-set by comparing  the audio, the reference transcripts and systems' output translations for both the cascade and direct models, as well as the ASR transcripts for the cascade one.
%
%
Some interesting examples of the identified errors are reported in Table~\ref{tab:audio-examples}.

\begin{table}[th!]
 \centering
 \small
 \tabcolsep2pt
\begin{tabular}{ll}
\hline
 \textsc{audio}  &  to the \textbf{er- euh} \textit{[disfluency]} Egyptian government  \\
 \textsc{c}      & der \textbf{eruptiven} \textit{[Eng. ``eruptive'']} Regierung ...\\ 
 
 \textsc{d}  & an die Regierung Ägyptens\\
\hline
 \textsc{audio}  &  dominated by \textbf{big, scary} guys,...  \\
 \textsc{c}      &   dominados por \textit{grandes} tipos \textit{aterradores} \\
 \textsc{d}  &   dominados por los chicos de \textbf{Big Kerry}  \\
\hline 
\textsc{audio}  &  I think, like \textbf{her},...  \\
  \textsc{c}      &  Penso che, come \textbf{qui} \textit{[Eng. ``here'']}, ...\\
 \textsc{d}  &  Penso che, come i \textbf{capelli} \textit{[Eng. ``hair'']},  ... \\
\hline
\end{tabular}
       \caption{Examples of audio understanding errors.} 

    \label{tab:audio-examples}
\end{table}

As shown in Table~\ref{tab:understanding}, 
audio understanding errors are 
quite common for both systems 
in all language pairs. However, both the number of errors and the number of sentences 
they affect
is significantly lower for the direct
one. 
We observed that this 
is the case especially for ``more difficult'' sentences, such as sentences with poor audio quality and overlapping or disfluent speech.

Though far from being conclusive (we acknowledge that, due to the ``opacity'' of direct models, their error counts might be slightly underestimated), 
this analysis seems to 
confirm 
the theoretical advantages of direct ST. 
This finding advocates for more thorough future 
investigations
on neural networks' interpretability, targeting
its empirical  verification on larger and diverse 
benchmarks.



\begin{table}[h]
    \centering
    \tabcolsep2.5pt
    \small
    \begin{tabular}{l|c|c|c||c|c||c|c}\hline
         & Both & C & D & C$_{tot}$ & D$_{tot}$ & C$_{sent}$ & D$_{sent}$  \\ \hline
        de & 51 & 96 & 52 & 147 & 103  & 117 & 91\\
        es & 82 & 108 & 66 & 190 & 148 & 150 & 127 \\
        it & 87 & 82 & 69 & 169  & 156 & 143 & 138\\ \hline
    \end{tabular}
    \caption{Audio understanding errors in the PE-set  and number of sentences containing at least one such error.}
    \label{tab:understanding}
\end{table}

\noindent \textbf{Prosody.}
Prosody is central to disambiguating utterances, as it reflects language elements 
which 
may not be encoded by grammar and vocabulary choices. While prosody is directly encoded by the direct system, it 
is lost in the unpunctuated input received by the MT component of a cascade. Besides few interrogative sentences, 
our annotators were able to isolate only a 
handful
of utterances 
whose prosodic markers 
result in different interpretations by the two models. 
%
Concerning  interrogatives, 
both systems 
managed to translate 
them correctly in most cases (24 for cascade and 25 for direct out of 31). This is not surprising given the syntactic structure of English questions, which is explicit and does not rely solely on prosody 
(e.g. compared to Italian).
%
In all other cases  (examples in Table~\ref{tab:prosody}), 
the direct model's higher sensitivity to prosody seems to give it an edge on cascade in disambiguating and correctly rendering the utterance meaning.
Also this finding calls for future inquiries
aimed to check the regularity 
of these 
differences  on larger datasets.



\begin{table}[h]

\small
\setlength\tabcolsep{1.3pt}
    \centering
    \begin{tabular}{ll}
    \hline
      src & nation states — governments doing the attacks \\
      C& Regierungen der Nationalstaaten  \\ 
      & \textit{[governments of nation states]} \\
      D& Nationen, Regierungen  \\
      & \textit{[nations, governments]} \\ 
      \hline
      src & 
      like the one we saw before, moving \\
      C& como el que vimos antes de moverse  \\
      & \textit{[like the one we saw before moving]}\\ 
      D& como el que hemos visto antes, moviéndose \\
      & \textit{[like the one we saw before, moving]}\\
      \hline
      src &  Photos like this: construction going on \\
      C& Foto come questa costruzione  
      \\ 
      & \textit{[Photos like this construction]} \\
      D& Foto come queste: costruzione  
      \\ 
      & \textit{[Photos like these: construction]} \\ 
      \hline
    \end{tabular}
        \caption{The two approaches dealing with prosody.}

    \label{tab:prosody}
\end{table}

%

\section{Linguistic Errors}
\label{sec:errors}
\subsection{Automatic Analysis}
\label{subsec:auto-errors}

For this analysis, we rely on the publicly available tool\footnote{\url{wit3.fbk.eu/2016-02}, details in Appendix \ref{appsec:tool}.} 
used  by \citet{Bentivogli2018} to analyse  what linguistic phenomena are best modeled by MT systems.
The tool exploits manual post-edits and HTER-based computations to detect and classify translation errors according to three linguistic categories: lexicon, morphology and word order. Table~\ref{tab:allErrors} presents their distribution.
%
%
%
%
%
%
%
%
%

%
As expected from the HTER scores 
in Table~\ref{tab:overall}, results vary across language pairs. On en-it, systems show pretty much the same number of errors, with a slight percentage gain 
(+1.1)
in favor of the cascade. For the other two pairs, differences are more marked and opposite, with an overall error reduction for the direct system  on en-es (-6.7)  and in favor of the cascade  on en-de (+6.7).

 

\COMMENT{
\begin{table*}[!h]
    \centering 
    \small
    \begin{tabular}{l|r|r|r||r|r|r||r|r|r}
       & \multicolumn{3}{|c|}{en-de} & \multicolumn{3}{|c|}{en-es} & \multicolumn{3}{|c}{en-it}\\
       \hline
     & Cascade & Direct & $\Delta \%$ & Cascade & Direct & $\Delta \%$ & Cascade & Direct & $\Delta \%$  \\
    \hline
    Lexical           & 2481    & 2560   & +3.18 &   2674 &	2497 &	-6.62  & 2264       & 2264   & 0.00  \\
    Morphology         & 468 &	536 &	+14.53 &   535  &	494 &	-7.66  & 433        &  470  & +8.55 \\
    Reordering        & 398	& 476 &	+19.60  & 308	 & 290	& -5.84  &  230        & 226    & -1.74   \\
 
    \hline
    Total                  & 3347	& 3572	& +6.72     &  3517 &	3281 &	-6.71 & 2927   & 2960  & +1.13\\

    \hline 
    \end{tabular}
    \caption{Distribution of error types for both ST approaches. The absolute number of errors (\#) is presented together with the percentage error reduction/increase of the  direct system with respect to the cascade one ({$\Delta \%$}). The morphology+reordering category includes those cases where both a morphological and reordering error occur on the same word.
    } 
    \label{tab:allErrors}
\end{table*}
}

\begin{table}[!t]
    \tabcolsep2pt
    \centering 
    \footnotesize
    \begin{tabular}{l||r|r|r||r|r|r||r|r|r}
       & \multicolumn{3}{c||}{en-de} & \multicolumn{3}{c||}{en-es} & \multicolumn{3}{c}{en-it}\\
       \cline{2-10}
     & \multicolumn{1}{c|}{C} & \multicolumn{1}{c|}{D} & $\Delta \%$ & \multicolumn{1}{c|}{C} & \multicolumn{1}{c|}{D} & $\Delta \%$ & \multicolumn{1}{c|}{C} & \multicolumn{1}{c|}{D} & $\Delta \%$ \\
    \hline
    L         & 2481    & 2560   & +3.2     & 2674 &	2497 &	-6.6 &  2264       & 2264   & 0.0\\
    M        & 468 &	536 &	+14.5 &  535  &	494 &	-7.7  &  433        &  470  & +8.6  \\
    R       & 398	& 476 &	+19.6  &   308	 & 290	& -5.8  &  230        & 226    & -1.7   \\
 
    \hline
                     & 3347	& 3572	& +6.7    &  3517 &	3281 &	-6.7  & 2927   & 2960  & +1.1 \\
    \hline 
        \end{tabular}
    \caption{Distribution of (L)exical, (M)orphological and (R)eordering 
    errors. 
    Absolute numbers are presented together with the percentage of reduction/increase of the  (D)irect system with respect to the (C)ascade 
    ({$\Delta \%$}). 
    } 
    \label{tab:allErrors}
\end{table}

%
%
%
Looking at 
the distribution of errors across categories, 
while for en-es the direct system is always better and the percentage reduction is homogeneously distributed, for en-de the better performance of the cascade 
is concentrated in the  morphology and word order categories. 
Since English and German are the most different languages in terms of morphology and word order, this result suggests that 
cascade systems still have an edge on the direct ones in their ability to handle morphology and word reordering.
This is further supported  by 
en-it: the only difference, in favor of the cascade, is 
indeed 
observed in the morphology category.

\subsection{Manual Analysis}
\label{subsec:man-errors}

Since lexical  errors represent
by  far  the  
 most frequent category for both approaches
in  all language pairs, we complement
the automatic analysis with a more fine-grained manual inspection, further distinguishing among lexical errors
due to missing words, 
extra words, or wrong lexical choice.\footnote{Various error taxonomies covering different levels of
granularity have been developed, and the distinction between these types of lexical errors is widely adopted, including the 
DQF-MQM framework -- \url{https://info.taus.net/dqf-mqm-error-typology-templ}}
%

The analysis was
carried out on 
subsets
of the 
PE-set, created
in such a way to be suitable for manual annotation. Namely, we
removed sentences for which the output of the two systems
is:
\textit{i)}  identical, \textit{ii)} judged correct by post-editors (HTER=0),
or \textit{iii)} 
too poor
to be reliably 
annotated for errors  (HTER$>$40\%).
%
%
%
The resulting
sets contain  207 sentences for en-de,  238 for en-es, 
and  
285 for en-it. 

This analysis reveals that, for all language pairs, wrong lexical choice is the most frequent 
error type
($\sim$65\% of lexical errors on average) followed by missing words ($\sim$30\%), 
and extra words ($\sim$5\%).

While
errors due to lexical choice and superfluous words vary across languages, we observe a systematic behavior with respect to missing words
(words that are present in the audio but are not translated). As we can see in Table~\ref{tab:missing}, 
direct systems lose more information from the source input than their cascade counterparts, in terms of both single words and 
contiguous word
sequences. 
It is particularly interesting to notice that also for en-es -- where the direct system is significantly stronger than the cascade -- the issue is still evident, although to a lesser extent.
Table~\ref{tab:examples} collects examples of the encountered lexical phenomena.




\begin{table}[h]
    \centering
    \small
    \begin{tabular}{l||c|c||c|c||c|c|c}

    & \multicolumn{2}{c||}{single } & \multicolumn{2}{c||}{word }  & \multicolumn{3}{c}{total }\\
    & \multicolumn{2}{c||}{words } & \multicolumn{2}{c||}{sequences }  & \multicolumn{3}{c}{\# words}\\
       \cline{2-8}
            & C & D & C & D & C & D & $\Delta \%$   \\ \hline
        de  & 25 & 34 & 6  &  10 & 42 & 58 & +38.10\\
        es & 26  &  40 & 10  & 11   & 59 & 68 & +15.25\\
        it  & 53  &  83 &  14 & 18  & 96 & 128 & +33.33 \\ \hline
    \end{tabular}
    \caption{Missing words  for (C)ascade and (D)irect systems. 
    Absolute numbers vary across languages as they reflect the different size of the annotated subsets.}
    \label{tab:missing}
\end{table}


\begin{table}[th!]
 \centering
 \small
 \setlength{\tabcolsep}{5pt}
\begin{tabular}{ll}
\hline
\textsc{audio}  &  ``That's fine'', \textbf{says} George,  \\
\textsc{c}      &  ``Das ist in Ordnung.'' \textbf{[ -- ]}  George,\\
\textsc{d}  & ``Das ist in Ordnung, \textbf{[ -- ]} George,'' \\
\hline
\textsc{audio}  &  \textbf{Well} after two years, ...  \\
\textsc{c}      &  \textit{Bueno}, después de dos años, ... \\
\textsc{d} & \textbf{[ -- ]} Después de dos años, ...\\
\hline
\textsc{audio}  &  My wife \textbf{and kids} and I, moved to  ...  \\
\textsc{c}      &  Io e mia moglie \textit{e i miei figli} ci siamo trasferiti...\\
\textsc{d}  & Io e mia moglie \textbf{[ -- ]} ci siamo trasferiti...  \\
 \hline
\end{tabular}
    \caption{Examples of missing words.} 
    \label{tab:examples}
\end{table}

 Finally, we  report that a non-negligible amount of missing words (between 10\% and 20\%) is represented by discourse markers, i.e. words or phrases used to connect and manage what is being said (e.g. ``you know'', ``well'', ``now''). Although this is a frequent phenomenon in speech, not translating discourse markers cannot be properly considered as an error, since markers \textit{i)} do not carry semantic information, and \textit{ii)} can be intentionally dropped in some use cases, such as in subtitling.

\section{Classifiers' Verdict}
\label{sec:classif}

So far, our inquiry has been entirely driven by predefined assumptions (the importance of  certain audio properties) and linguistic criteria (the focus on specific error types).  This \textit{top-down} approach, however, might 
fail to disclose important differences, which  were not specifically sought after when analysing the two paradigms. This consideration motivates the adoption of the complementary \textit{bottom-up} approach that concludes our comparative study by answering the question: is the output of cascade and direct systems distinguishable? Understanding if and why discriminating between the two is possible would not only suggest new issues to  look at. It would also highlight possible output regularities that, despite the similar overall performance,  make one paradigm preferable over the other in specific application scenarios. To this aim, we set up a classification experiment, comparing the ability of humans to correctly identify the output of the two systems with the performance of an automatic text classifier. 

\subsection{Human Classification}
\label{subsec:humanclass}

After getting acquainted with systems' output through the previous manual analyses, our assessors 
were 
instructed to perform a classification task.
The classification had to be performed on 10 blocks of items 
comprising 
a set of 
unseen 
English
contiguous
sentences (gold 
transcripts)  from the  MuST-C \textit{Common} test set, 
and two sets of 
anonymized
translations, 
one produced by 
the cascade and one by the direct model. 
For each block, 
the assessors had to 
assign 
each set of translations to the correct system, or label them as indistinguishable.
%
%
To investigate whether more context helps in the assignment, we set up two experiments with respectively 10 and 20 
contiguous sentences per block.

\begin{table}[ht]
    \centering 
    \small
    \begin{tabular}{l|r|r||r|r||r|r}
      & \multicolumn{2}{c||}{en-de} & \multicolumn{2}{c||}{en-es} & \multicolumn{2}{c}{en-it}\\
      \hline
    \# of sentences & 10  & 20 &  10 & 20 & 10 & 20  \\
    \hline
    Correct             &  7	& 6 &  4  & 4  & 4    & 3 \\
    Wrong               &  2    & 2 &  2  & 3  & 1	  &	2  \\
    Indistinguishable   &  1    & 2 &  4  & 3  & 5	  &	5  \\
    \hline
    Total  \# of blocks         & 10   & 10  & 10   &  10 &	10 &	10 \\
    \hline 
    \end{tabular}
    \caption{Results of human classification.} 
    \label{tab:HumClassification}
\end{table}

The results 
in Table~\ref{tab:HumClassification} show that en-es and en-it systems are 
not distinguishable, since only a maximum of 4 blocks out of 10 
were correctly classified, while most en-de blocks were correctly classified. 
%
According to the en-de assessor, this is due to the fact that the structure of the sentences generated by the direct system is very similar to that of the corresponding English sources. 
This characteristic stands out in German, which differs from English in terms of word order more than Italian and Spanish.
This type of behavior does not necessarily imply the presence of errors but, like a fingerprint, makes the en-de direct system more recognizable by a human.
Furthermore, being sub-optimal for German, this structure can cause preferential edits by the post-editors, which would be in line with the concentration of errors in the word order category observed in Table ~\ref{tab:allErrors}  (+19.6\%).

Assessing the importance of context, the ability of humans to distinguish the systems
does not improve when passing from 10 to 20 sentences per block. This 
suggests that the behavioral differences between cascade and direct systems
are so subtle 
that, on larger samples, they mix up and balance
making their fingerprints less traceable. 

\subsection{Automatic Classification}
\label{subsec:autoclass}

As a complement to the human classification experiment, 
we
check whether an automatic tool is able to accomplish 
a similar 
task.
Our classifier combines \textit{n}-gram language models with the Naive Bayes algorithm, as proposed in~\citep{peng-ecir2003}. 
We trained two 5-gram 
models, respectively using translations by the cascade and the direct systems.
At classification time, given a translated text,
the classifier computes the perplexity of the two models and assigns the \textit{cascade} or \textit{direct} label based on the model with the lowest perplexity.
%
%
Also these experiments were carried out
on the MuST-C \textit{Common} set.
The classifier was tested via k-fold cross-validation, 
for different values of k -- i.e. different sizes of text to classify. 

\begin{figure}[h]
    \hspace{-3mm}\includegraphics[width=0.5\textwidth]{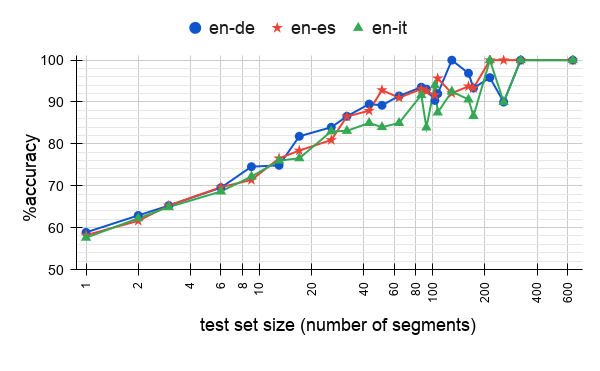}
    \vspace*{-5mm}
        \caption{Results of automatic classification for different sizes of system output 
blocks (1-600 sentences).}
    \label{fig:en_it_classification}
\end{figure}

As shown in Figure \ref{fig:en_it_classification}, contrary to humans, the more data the classifier receives,
the higher its accuracy in  discriminating between systems. 
Already at a size of 20 sentences, accuracy is always 
$\sim$80\%.
%
%
This suggests that 
systems have their own 
``language'', a fluency-related fingerprint.

To check this finding, we measured outputs'
\textit{lexical diversity}  in terms of 
moving average Type-Token Ratio -- maTTR \cite{covington-2010-mattr} -- and 
with the Measure of Textual Lexical Diversity (MTLD) by \citet{McCarthy-2010-MTLD}. 

Table \ref{tab:lexdiv} shows that the cascade output exhibits higher lexical 
diversity on all 
languages, with smaller differences on en-de and en-es compared to en-it.
%
%
A plausible conclusion is that the cascade produces richer output, whose variety
does not necessarily result in better translations nor is appreciated by humans.
Indeed, 
annotators 
were able to correctly distinguish the output 
only for en-de, where 
lexical diversity 
is similar 
(see $\S$\ref{subsec:humanclass}).



\begin{table}[ht]
    \setlength{\tabcolsep}{5pt}
    \centering 
    \small
    \begin{tabular}{l|r|r|r|r|r|r}
      & \multicolumn{2}{c|}{en-de} & \multicolumn{2}{c|}{en-es} & \multicolumn{2}{c}{en-it}\\
      \cline{2-7}
     & {\scriptsize maTTR}  & {\scriptsize MTLD} & {\scriptsize maTTR}  & {\scriptsize MTLD} & {\scriptsize maTTR}  & {\scriptsize MTLD}  \\
    \hline
    R &  73.11    & 97.02   &  69.81  & 77.19  & 74.50	  &	 109.79  \\
    C   &  71.84    & 83.64   &  68.42  & 67.68  & 73.20	  &	 97.82  \\
    D    &  71.45    & 83.27   &  67.99  & 65.59  & 72.60	  &	 90.78  \\
    \hline
    \end{tabular}
    \caption{Lexical diversity of the human (R)eference,  (C)ascade and (D)irect outputs.} 
    \label{tab:lexdiv}
\end{table}

\section{Conclusion and Final Remarks}
%
There is a time when the possible transition from consolidated technological frameworks to new emerging paradigms depends on answering fundamental questions about their  potential, 
strengths and weaknesses. A time when technology developers are faced with  the choice of where to direct their future investments. Five years after its appearance on the scene, the direct approach to ST confronts the community with 
similar questions in relation to the traditional cascade paradigm that it aims to overtake.
Our 
investigation 
showed that, in spite of the known data paucity conditions still penalizing the direct approach, the two technologies now perform substantially on par.
Subtle differences in their behavior 
exist: overall performance being equal, 
the cascade 
still seems to have
an edge in terms of morphology, word ordering and lexical diversity, which is balanced by the advantages of direct models in audio understanding and in capturing prosody. However, they do not seem sufficient 
and consistent enough across languages 
to
make the output of the two approaches easily distinguishable, nor to make one model preferable to the other. Back to our title, they no longer make a difference.
%
%

We 
are aware that 
the generalizability of these results depends on several 
factors 
%
%
%
such as the considered languages,
systems and benchmarks, as well as the human workforce deployed for the inquiry.
Here, with the help of professionals, we proposed multi-faceted quantitative and qualitative analyses, 
run on the output of state-of-the-art systems on three language pairs -- though, by now, covering only the most-explored and data-favorable condition, which has English as source.
Although our findings hold for a specific scenario, in which free data 
were at our disposal (and to which we contribute back by releasing high-quality post-edits), they might not be generalizable to other
(e.g. difficult, distant) languages  and other (e.g. highly specialized) domains.  Nevertheless, we present them as a
timely contribution towards answering a burning question within the ST community.

\section*{Acknowledgements}

The creation of the post-edits used in this work was funded by the European Association for Machine Translation (EAMT) through its 2020 Sponsorship of Activities programme. The computational costs were covered by the ``End-to-end Spoken Language Translation in Rich Data Conditions'' project,\footnote{\url{https://ict.fbk.eu/units-hlt-mt-e2eslt/}} which was financially supported by an Amazon AWS ML Grant.

\bibliographystyle{acl_natbib}
\bibliography{fbk,anthology,acl2021}

\appendix

\section{Systems' Description}
\label{appsec:systems}

In this section we describe
the ST models created for our study (see Section 3.1
). All the details about the different trainings are given below, while the validation set was common to all trainings, since we used the MuST-C \textit{dev set}.

The source code for the ASR and the direct ST models is available at: 
\url{https://github.com/mgaido91/FBK-fairseq-ST}.

The source code for the MT component of the cascade model can be found at: 
\url{https://github.com/modernmt/modernmt}.



\subsection{Cascade approach}
\label{sssec:cascade}

The  Cascade system is composed of a pipeline of automatic speech recognition (ASR) and machine translation (MT) models.

The \textbf{ASR} model is a slightly revisited version \citep{gaido-etal-2020-end} of the S-Transformer \citep{digangi2019adapting}, where the two 2D self-attention
layers are replaced with two Transformer encoder layers (for a total of 8 layers), while the decoder is the same (with 6 layers).
Hence, the model  processes the input with two 3x3 2D CNNs (having 64 filters), whose output is first projected into a higher-dimensional space and then  summed with positional embeddings before being fed to the Transformer encoder layers; Transformer encoder layers use logarithmic distance penalty. The attention mechanism consists of 8 attention heads. The dimensionality of input and output is 512, while the inner-layers have dimensionality 2048. The resulting number of parameters is 63M. 

The ASR model was trained with the goal of achieving state-of-the-art performance.
To this aim, we relied on two data augmentation
techniques that were shown to yield competitive models at the IWSLT-2020 evaluation campaign \cite{ansari-etal-2020-findings}, namely:
\textit{i)} SpecAugment~\cite{Park_2019} applied with probability 0.5 by masking two bands on the frequency axis (with 13 as maximum mask length) and two on the time axis (with 20 as maximum mask length), and \textit{ii)} time stretch~\cite{nguyen2019improving} with probability of 0.3 and stretching factor sampled uniformly for each utterance between 0.8 and 1.25. 
The ASR model was trained on 1.25M 
utterance-transcript pairs coming from the ASR corpora Librispeech \cite{librispeech}, Mozilla Common Voice,\footnote{\url{https://voice.mozilla.org/}} How2 \cite{sanabria18how2}, TEDLIUM-v3 \cite{Hernandez_2018}, as well as the ST corpora Europarl-ST \cite{europarlst} and MuST-C \cite{MuST-Cjournal}.\footnote{For English-German, the ST corpora include also the Speech-Translation TED corpus
provided in the IWSLT offline-speech-translation task: \url{http://iwslt.org/doku.php?id=offline\_speech\_translation}}
We filtered out all pairs whose utterance was longer than 20 seconds.
The audio input was preprocessed with XNMT\footnote{\url{https://github.com/neulab/xnmt}} \cite{neubig18xnmt} to extract 40 features per time frame (with 25ms windows and 10ms sliding) and 
per-speaker normalization was applied.
The text was preprocessed by normalizing punctuation and de-escaping special characters, and was tokenized with Moses.\footnote{\url{https://github.com/moses-smt/mosesdecoder}} Then it was encoded with a BPE~\cite{sennrich2015neural} code learnt on the OPUS data\footnote{\label{foot:opus}\url{http://opus.nlpl.eu}} using 8k merge rules. 

The \textbf{MT} component is built on the ModernMT framework\footnote{\url{https://github.com/modernmt/modernmt}} which features machine translation implementing the Transformer architecture. We trained either a Base (en-it) or a Big (en-\{de,es\}) Transformer model~\cite{transformer} with 6 blocks in the encoder and 6 in the decoder, 512/1024 as input size, the same as output size, 2048/4096 as inner dimension and 8/16 attention heads.  The total number of parameters is about 61M for the Base model, 210M for the Big models. 

As regards pre-processing, for all the three language directions we used the internal ModernMT procedures.

In training, 
models are optimized with Adam using $\beta_1$=0.9, $\beta_2$=0.98; the learning rate is linearly increased during the warmup (8k iterations) up to the maximum value ($5\times 10^{-4}$), after that it follows an inverse square root decay; dropout is set to 0.3.  Minibatches consist of 3072 tokens and update frequency is set to 4; the total number of iterations is 200k; the last 10 saved checkpoints (one out of 1k iterations) are averaged.  The model uses label smoothing with a uniform prior distribution (0.1) over the vocabulary; source and target languages share a BPE
vocabulary of 32k sub-words.

\begin{table}[ht]
 \centering  
  \setlength{\tabcolsep}{5pt}
\begin{tabular}{|l|c|c|c|}
\cline{2-4}
\multicolumn{1}{c|}{} & \#segments & \#en words & \#trg words \\
 \hline
en-de & 58.2M & 776.4M & 723.3M\\
en-es & 70.1M & 972.5M & 1024.9M\\
en-it & 67.9M & 792.6M & 770.2M \\
\hline
\end{tabular}
    \caption{Statistics of the parallel training sets collected from the OPUS repository for the three languages pairs.} 
    \label{tab:MTtrainStat}
\end{table}

The training data, whose statistics are reported in Table~\ref{tab:MTtrainStat}, are collected from the OPUS repository.
For English-Italian, they resulted in almost 70M segment pairs and about 800M English words; after deduplication and the internal ModernMT cleaning, the actual training data is reduced to 45M pairs and 550M English words. For English-\{German,Spanish\} pairs, the OPUS data were filtered through well-known data selection methods \cite{axelrod-etal-2011-domain} using a general-domain seed; the resulting training data consist of, respectively, 17M and 19M segment pairs, for 270M and 330M English words. Trainings were performed on RTX 2080 Ti GPUs; for English-Italian, it was run on 7 GPUs and lasted 3 days, while for each of the other two directions, on a single GPU, it took 6 days.

The three models are then fine-tuned on MuST-C training data ($\sim$250K pairs, 4-5M English words) by continuing the training for 4k iterations on the adaptation data, with a learning rate reduced by a factor of 5.  To mitigate error propagation and make the MT system more robust to ASR errors, similarly to \cite{digangi2019robust} fine tuning is run on the concatenation of human and automatic transcripts of MuST-C, both paired with manual translations.

\subsection{Direct approach}
\label{sssec:e2e}


Our direct model \citep{gaido-etal-2020-end} uses the same architecture of the English ASR model described in 
$\S$\ref{sssec:cascade}, but it has 11 Transformer encoder layers (instead of 8) and 4 Transformer decoder layers (instead of 6) for a total of 64M parameters.
The ST model's encoder is initialized with the encoder of the ASR model \cite{bansal-etal-2019-pre}, with the missing layers  initialized randomly. The ST decoder is also initialized randomly.


The training settings and the data augmentation methods employed for the direct ST model are the same described in Section \ref{sssec:cascade} for the ASR component of the cascade system. In addition, we performed synthetic data generation, by automatically translating the English transcripts of the ASR training corpora \cite{jia2018leveraging}. 
Furthermore, we  transfer knowledge
from MT through knowledge distillation~\cite{hinton2015distilling}.
Knowledge distillation is performed
from a \textit{teacher} MT model by optimizing the KL divergence between the distributions produced by the teacher and the \textit{student} ST model being trained \cite{liu2019endtoend}.
The teacher MT model is trained on the OPUS datasets  \cite{opus} and is a plain transformer with 16 attention heads and 1024 features in encoder/decoder embeddings, resulting into 212M parameters.

The direct ST model is trained in two consecutive steps. First, it is optimized using KD.
Then, the resulting model is fine-tuned on label-smoothed cross entropy \cite{szegedy2016rethinking}.
The training set is composed of the same corpora  used for the ASR model, more precisely: \textit{i)} the ST corpora and \textit{ii)} the synthetic datasets derived from the ASR corpora. 

The ST model is fed with the input utterance and a token representing the \textit{type}
of the target data, which can be: \textit{i)} human reference translations (for the ST corpora), or \textit{ii)} translations generated by the MT model fed with true case transcriptions with punctuation, and \textit{iii)} translations generated by the MT model fed with lower-cased transcriptions without punctuation (for the ASR corpora).
At inference time, the token ``human reference'' is always used to generate the translations. The token is added to the features extracted from the audio before they are passed to the encoder \cite{digangi2019one}.

All trainings were performed on 8 K80 GPUs. The training of each direct model lasted 10 days, while the ASR and MT pre-trainings 6 days each.

The source code\footnote{\url{https://github.com/mgaido91/FBK-fairseq-ST}} implemented to build these models is based on Fairseq~\citep{ott-etal-2019-fairseq}.


\section{Post-Editing Guidelines}
\label{appsec:pe_guidelines}
In this task you are presented with (i) 550 audio segments that are recordings of portions of different English TED Talks, (ii) their transcripts, and (iii) corresponding automatic translations. 

Starting from the original audio recording and its corresponding transcript (done by TED volunteer translators), you are asked to post-edit each given automatic translation by applying the minimal edits required to transform the system output into a fluent sentence with the same meaning as the audio/transcript. 

While post-editing, remember the following guidelines:
\begin{itemize}
\item We noticed that some audio player software applications cut the beginning or the end of the audio segments. If you notice some audio-transcript out-of-sync, please try another audio player or inform us about the problem.
\item The audio should be your first source of information, while transcripts are given for your convenience.  It could happen that the transcript is not faithful to the spoken original: in these cases you should not consider the transcript and refer to the audio only.
\item Some transcripts contain the name or initials of the speaker (typically followed by colons). Please don't add this information into the sentence you are post-editing. In general, don't include in your post-edit any text that is not present in the audio (e.g. explanation of acronyms, disambiguation of pronouns), even though this information could ease the understanding of the sentence.
\item The post-edited sentence is intended as a translation of spoken language. Also, depending on the style of the source language talk, you can use the corresponding style in the target language (e.g. if the talk uses a friendly/colloquial style you can use informal words too).
\item The focus is the correctness of the single sentence within the given context, not the consistency of a group of sentences.  Hence, surrounding segments should be used to understand the context but not to enforce consistency on the use of terms. In particular, different but correct translations of terms across segments should not be corrected.
\end{itemize}

\section{Tool for Automatic Error Classification }
\label{appsec:tool}



The tool used for the automatic analysis of linguistic errors (Section 6.1) is downloadable at \url{wit3.fbk.eu/2016-02}. It is a modified version of the \textit{tercom} script,~\footnote{\url{www.cs.umd.edu/~snover/tercom}} 
which requires the lemmatized versions of both systems' outputs and post-edits. To  lemmatize the data we used the \textit{TreeTagger}.\footnote{\url{www.cis.uni-muenchen.de/~schmid/tools/TreeTagger}} 

\end{document}


\maketitle

\section{Systems' Description}

In this section we give details about the models created for our study (Section 3.1\footnote{Section of the main paper.\label{MP}}). All the details about the different trainings are listed below, dividing them into the models composing the cascade solutions and those employed to produce the direct systems.

The validation set is common to all trainings, since we used the MuST-C \textit{dev set}.

All systems for all language directions were evaluated on two test sets: i) the MuST-C \textit{Common} test set and ii) the PE-set, which was specifically created for this study \lb{and will be publicly released as a further contribution of our work}.

\subsection{Cascade System}
\label{sssec:cascade}

The  Cascade system is composed of a pipeline of automatic speech recognition (ASR) and machine translation (MT) models.

\newcommand{\COMMENT}[1]{}
\COMMENT{
The ASR component is based on the KALDI toolkit \cite{povey2011kaldi}, featuring a time-delay neural network and lattice-free maximum mutual information discriminative sequence-training ~\cite{povey2016}. The audio data for acoustic modeling include the clean portion of LibriSpeech \cite{librispeech} ($\sim$460h) and a variable subset of the MuST-C training set ($\sim$450h), from which 40 MFCCs per time frame were extracted; a MaxEnt  language model~\cite{Alumae2010} is estimated from the corresponding transcripts ($\sim$7M words). 
}

The \textbf{ASR} model is a slightly revisited version of the S-Transformer
\cite{digangi2019adapting}
where the two 2D self-attention
layers are replaced with two Transformer encoder layers (for a total of 8 layers), while the decoder is the same (with 6 layers).
Hence, the model  processes the input with two 3x3 2D CNNs (having 64 filters), whose output is first projected into a higher-dimensional space and then  summed with positional embeddings before being fed to the Transformer encoder layers; Transformer encoder layers use logarithmic distance penalty. The attention mechanism consists of 8 attention heads. The dimensionality of input and output is 512, while the inner-layers have dimensionality 2048. The resulting number of parameters is 63M. The data augmentation methods employed and the training settings are the same described in Section \ref{sssec:e2e} for the direct systems.
Our ASR model was trained on 1.25M utterance-sentence pairs coming from Librispeech \cite{librispeech}, Mozilla Common Voice,\footnote{\url{https://voice.mozilla.org/}} How2 \cite{sanabria18how2}, TEDLIUM-v3 \cite{Hernandez_2018}, the ST corpora -- Europarl-ST \cite{europarlst} and MuST-C \cite{MuST-Cjournal}.\footnote{For English-German, the ST corpora include also the Speech-Translation TED corpus
provided in the IWSLT offline-speech-translation task: \url{http://iwslt.org/doku.php?id=offline\_speech\_translation}}
We filtered the pairs whose utterance was longer than 20s. The input text was preprocessed by normalizing punctuation, de-escaping special characters and tokenized with Moses.\footnote{\url{https://github.com/moses-smt/mosesdecoder}} Then it is encoded with a BPE~\cite{sennrich2015neural} code learnt on the MT data using 8k merge rules. The audio, instead, was preprocessed with XNMT\footnote{\url{https://github.com/neulab/xnmt}} \cite{neubig18xnmt} to extract 40 features per time frame (with 25ms windows and 10ms sliding) and apply per-speaker normalization.

The \textbf{MT} component is built on the ModernMT framework\footnote{\url{https://github.com/modernmt/modernmt}} which features machine translation implementing the Transformer architecture. We trained either a Base (en-it) or a Big (en-\{de,es\}) Transformer model~\cite{transformer} with 6 blocks in the encoder and 6 in the decoder, 512/1024 as input size, the same as output size, 2048/4096 as inner dimension and 8/16 attention heads.  The total number of parameters is about 61M for the Base model, 210M for the Big models. 

As regards pre-processing, for all the three language directions we used the internal ModernMT procedures.

In training, 
models are optimized with Adam using $\beta_1$=0.9, $\beta_2$=0.98; the learning rate is linearly increased during the warmup (8k iterations) up to the maximum value ($5\times 10^{-4}$), after that it follows an inverse square root decay; dropout is set to 0.3.  Minibatches consist of 3072 tokens and update frequency is set to 4; the total number of iterations is 200k; the last 10 saved checkpoints (one out of 1k iterations) are averaged.  The model uses label smoothing with a uniform prior distribution (0.1) over the vocabulary; source and target languages share a BPE~\cite{sennrich2015neural} vocabulary of 32k sub-words.

\begin{table}[ht]
 \centering  
  \setlength{\tabcolsep}{5pt}
\begin{tabular}{|l|c|c|c|}
\cline{2-4}
\multicolumn{1}{c|}{} & \#segments & \#en words & \#trg words \\
 \hline
en-de & 58.2M & 776.4M & 723.3M\\
en-es & 70.1M & 972.5M & 1024.9M\\
en-it & 67.9M & 792.6M & 770.2M \\
\hline
\end{tabular}
    \caption{Statistics of the parallel training sets collected from the OPUS repository for the three languages pairs.} 
    \label{tab:MTtrainStat}
\end{table}

The training data, whose statistics are reported in Table~\ref{tab:MTtrainStat}, are collected from the OPUS repository.\footnote{\url{http://opus.nlpl.eu}} For English-Italian, they resulted in almost 70M segment pairs and about 800M English words; after deduplication and the internal ModernMT cleaning, the actual training data is reduced to 45M pairs and 550M English words. For English-\{German,Spanish\} pairs, the OPUS data were filtered through well-known data selection methods \cite{axelrod-etal-2011-domain} using a general-domain seed; the resulting training data consist of, respectively, 17M and 19M segment pairs, for 270M and 330M English words. Trainings were performed on RTX 2080 Ti GPUs; for English-Italian, it was run on 7 GPUs and lasted 3 days, while for each of the other two directions, on a single GPU, it took 6 days.

The three models are then fine-tuned on MuST-C training data ($\sim$250K pairs, 4-5M English words) by continuing the training for 4k iterations on the adaptation data, with a learning rate reduced by a factor of 5.  To mitigate error propagation and make the MT system more robust to ASR errors, similarly to \cite{digangi2019robust} fine tuning is run on the concatenation of human and automatic transcripts of MuST-C, both paired with manual translations.

\subsection{Direct approach}
\label{sssec:e2e}


Our direct model uses the same architecture of the ASR model described in 
$\S$\ref{sssec:cascade}, but it has 11 Transformer encoder layers and 4 Transformer decoder layers for a total of 64M parameters.
It is trained with the goal of achieving state-of-the-art performance on the ST task.
To this aim, we rely on  data augmentation  and knowledge transfer techniques that were shown to yield competitive models at the IWSLT-2020 evaluation campaign \cite{ansari-etal-2020-findings}.
In particular, we use
three data augmentation methods: \textit{i)} SpecAugment~\cite{Park_2019} applied with probability 0.5 by masking two bands on the frequency axis (with 13 as maximum mask length) and two on the time axis (with 20 as maximum mask length), \textit{ii)} time stretch~\cite{nguyen2019improving} with probability of 0.3 and stretching factor sampled uniformly for each utterance between 0.8 and 1.25, and  \textit{ii)} synthetic data generation from ASR corpora \cite{jia2018leveraging}. In addition, we  transfer knowledge both from ASR and MT through component initialization and knowledge distillation~\cite{hinton2015distilling}.

The ST model's encoder is initialized with the encoder of our English ASR model (see 
$\S$\ref{sssec:cascade}) \cite{bansal-etal-2019-pre}; the missing layers are initialized randomly, as well as the decoder.
Knowledge distillation (KD) is performed
from a \textit{teacher} MT model by optimizing the KL divergence
between the distribution produced by the teacher and 
by the \textit{student} ST model being trained \cite{liu2019endtoend}.
The MT model is trained on the OPUS datasets  \cite{opus} and is a plain transformer with 16 attention heads and 1024 features in encoder/decoder embeddings, resulting into 212M parameters.

The model is trained in two consecutive steps. First, it is optimized using KD.
Then, the resulting model is fine-tuned on label-smoothed cross entropy \cite{szegedy2016rethinking}.
The training set is composed of the same corpora  used for the ASR model.
The ST model is fed with the input utterance and a token representing the \textit{type}
of the target data, which can be: \textit{i)} the ground truth (for the ST corpora having a target reference),
\textit{ii)} generated by the MT model fed with true case transcriptions with punctuation,
and \textit{iii)} generated by the MT model fed with lower-cased transcriptions without punctuation.
At inference time, the first token is always used to generate the translations.
The token is added to the features extracted from the audio before they are passed to the encoder \cite{digangi2019one}.

The source code implemented to build these models is based on Fairseq~\citep{ott-etal-2019-fairseq}, and will be released open-source upon acceptance of the paper.

 All trainings were performed on 8 K80 GPUs. The training of a direct model lasted 10 days, while the ASR and MT pre-trainings 6 days each.

\section{Evaluation Metrics and Error Annotation Tool}

Automatic evaluation (Section 4\textsuperscript{\ref{MP}}) was carried out using the sacreBLEU and TER metrics.  The version signature of SacreBLEU\footnote{\url{https://github.com/mjpost/sacrebleu/}} is: BLEU+c.mixed+\#.1+s.exp+tok.13a+v.1.4.3, while we used the \textit{tercom} implementation of TER\footnote{\url{www.cs.umd.edu/~snover/tercom}} 
with default parameters.

The tool used for the automatic analysis of linguistic errors (Section 6.1\textsuperscript{\ref{MP}}) is downloadable at \url{wit3.fbk.eu/2016-02}. It is a modified version of the \textit{tercom} script  
requiring the lemmatized versions of both systems' outputs and post-edits. To  lemmatize the data we used the \textit{TreeTagger}.\footnote{\url{www.cis.uni-muenchen.de/~schmid/tools/TreeTagger}} 

\section{Post-editing Guidelines}
In this task you are presented with (i) 550 audio segments that are recordings of portions of different English TED Talks, (ii) their transcripts, and (iii) corresponding automatic translations. 

Starting from the original audio recording and its corresponding transcript (done by TED volunteer translators), you are asked to post-edit each given automatic translation by applying the minimal edits required to transform the system output into a fluent sentence with the same meaning as the audio/transcript. 

While post-editing, remember the following guidelines:
\begin{itemize}
\item We noticed that some audio player software applications cut the beginning or the end of the audio segments. If you notice some audio-transcript out-of-sync, please try another audio player or inform us about the problem.
\item The audio should be your first source of information, while transcripts are given for your convenience.  It could happen that the transcript is not faithful to the spoken original: in these cases you should not consider the transcript and refer to the audio only.
\item Some transcripts contain the name or initials of the speaker (typically followed by colons). Please don't add this information into the sentence you are post-editing. In general, don't include in your post-edit any text that is not present in the audio (e.g. explanation of acronyms, disambiguation of pronouns), even though this information could ease the understanding of the sentence.
\item The post-edited sentence is intended as a translation of spoken language. Also, depending on the style of the source language talk, you can use the corresponding style in the target language (e.g. if the talk uses a friendly/colloquial style you can use informal words too).
\item The focus is the correctness of the single sentence within the given context, NOT the consistency of a group of sentences.  Hence, surrounding segments should be used to understand the context but NOT to enforce consistency on the use of terms. In particular, different but correct translations of terms across segments should not be corrected.
\end{itemize}

\bibliographystyle{acl_natbib}
\bibliography{fbk,anthology,acl2021}
